\documentclass[letterpaper, 10 pt, conference]{ieeeconf}  

\IEEEoverridecommandlockouts                              

\usepackage{amsmath} 
\usepackage{amssymb}  
\usepackage[pdftex]{graphicx}
\usepackage{multirow}
\usepackage{amsfonts}
\usepackage{float}
\usepackage{pifont}
\usepackage{subcaption}
\usepackage{stfloats}
\usepackage{graphicx,adjustbox}
\usepackage{accents}
\usepackage{cite}
\usepackage{bm}
\usepackage{etex}
\usepackage{accents}

\usepackage{subcaption} 

\usepackage{tabularx}

\title{\LARGE \bf
Scene Exploration by Vision-Language Models
}

\author{Venkatesh Sripada, Samuel Carter, Frank Guerin, Amir Ghalamzan
\thanks{{$^{1}$All authors are with the University of Surrey, UK}}
}

\begin{document}

\maketitle
\thispagestyle{empty}
\pagestyle{empty}

\begin{abstract}
Active perception enables robots to dynamically gather information by adjusting their viewpoints, a crucial capability for interacting with complex, partially observable environments. In this paper, we present AP-VLM, a novel framework that combines active perception with a Vision-Language Model (VLM) to guide robotic exploration and answer semantic queries. Using a 3D virtual grid overlaid on the scene and orientation adjustments, AP-VLM allows a robotic manipulator to intelligently select optimal viewpoints and orientations to resolve challenging tasks, such as identifying objects in occluded or inclined positions. We evaluate our system on two robotic platforms: a 7-DOF Franka Panda and a 6-DOF UR5, across various scenes with differing object configurations. Our results demonstrate that AP-VLM significantly outperforms passive perception methods and baseline models, including Toward Grounded Common Sense Reasoning (TGCSR), particularly in scenarios where fixed camera views are inadequate. The adaptability of AP-VLM in real-world settings shows promise for enhancing robotic systems' understanding of complex environments, bridging the gap between high-level semantic reasoning and low-level control.
\end{abstract}

\section{Introduction}

Recent advancements in large language models (LLMs) and vision-language models (VLMs) have enabled the seamless integration of high-level semantic reasoning with low-level robotic control. These models excel at providing visual and semantic insights into a scene and can be directed through natural language to perform complex reasoning. By transforming abstract objectives into actionable sequences, LLMs offer valuable support for robotic systems that require semantic understanding to navigate and manipulate their environments.

A key challenge in robotic perception is actively gathering information to improve task performance, especially in cases where passive perception is insufficient. This is crucial when robots must interact with partially observable or occluded objects. For instance, identifying the contents of a mug that is not upright may require the robot to actively change its perspective. In such cases, coordinating perception and action is essential for overcoming epistemic uncertainties and accomplishing the task.

In order to bring the perception abilities of VLMs to the problem of active perception for robots, we need to bridge the gap between the normal language space of VLM output and the precise references in space needed for robotics. Inspired by recent works that make use of visual prompting \cite{saycan2022arxiv,liang2023code}, i.e., annotating the images input to the VLM, we overlay a 3D grid on the input scenes, this then allows the VLM to give the robot precise instructions to move to more advantageous viewing positions.

In this paper, we present a novel zero-shot framework for active perception in robotics, using a 6DOF UR5 and a 7-DOF Franka Panda robotic manipulator equipped with an in-hand camera. The robot explores the scene to answer specific queries posed by a VLM. Initially, it captures an image from a home configuration, annotated with a virtual grid anchored to a reference marker calibrated to the robot's base frame. When queried, such as determining the contents of an overturned mug, the VLM suggests the next optimal viewpoint. The robot then moves to this position to collect better visual data. This process repeats iteratively until the VLM can confidently respond.

\begin{figure}[tb!]
    \centering
    \begin{subfigure}[b]{0.48\linewidth}
        \centering
        {\includegraphics[width=\linewidth,trim={11cm 0cm 14cm 0cm},clip]{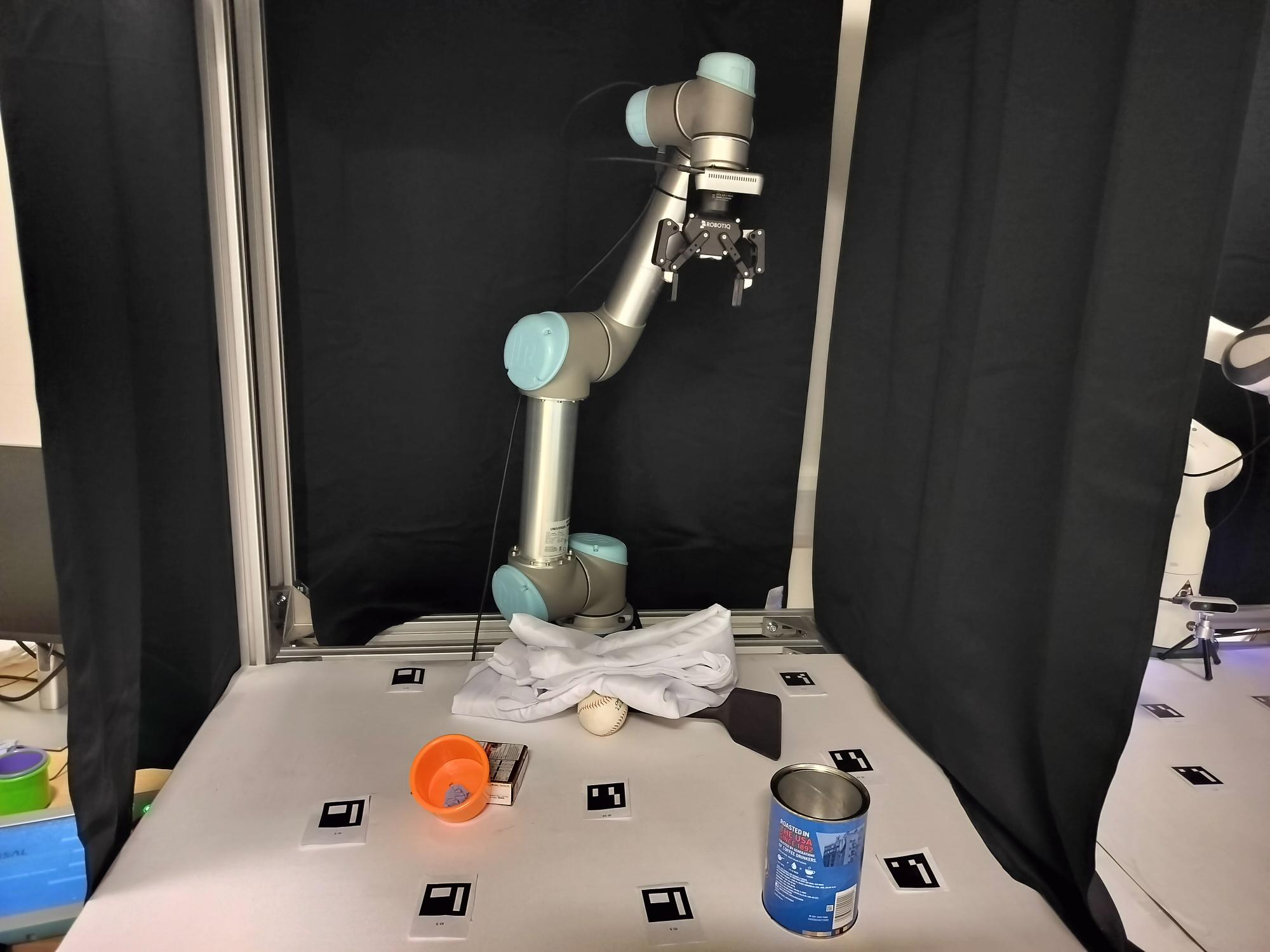}}
        \label{fig:scene_2}      
        \vspace{-.5cm}
    \end{subfigure}
    \hfill
    \begin{subfigure}[b]{0.48\linewidth}
        \centering
        {\includegraphics[width=\linewidth]{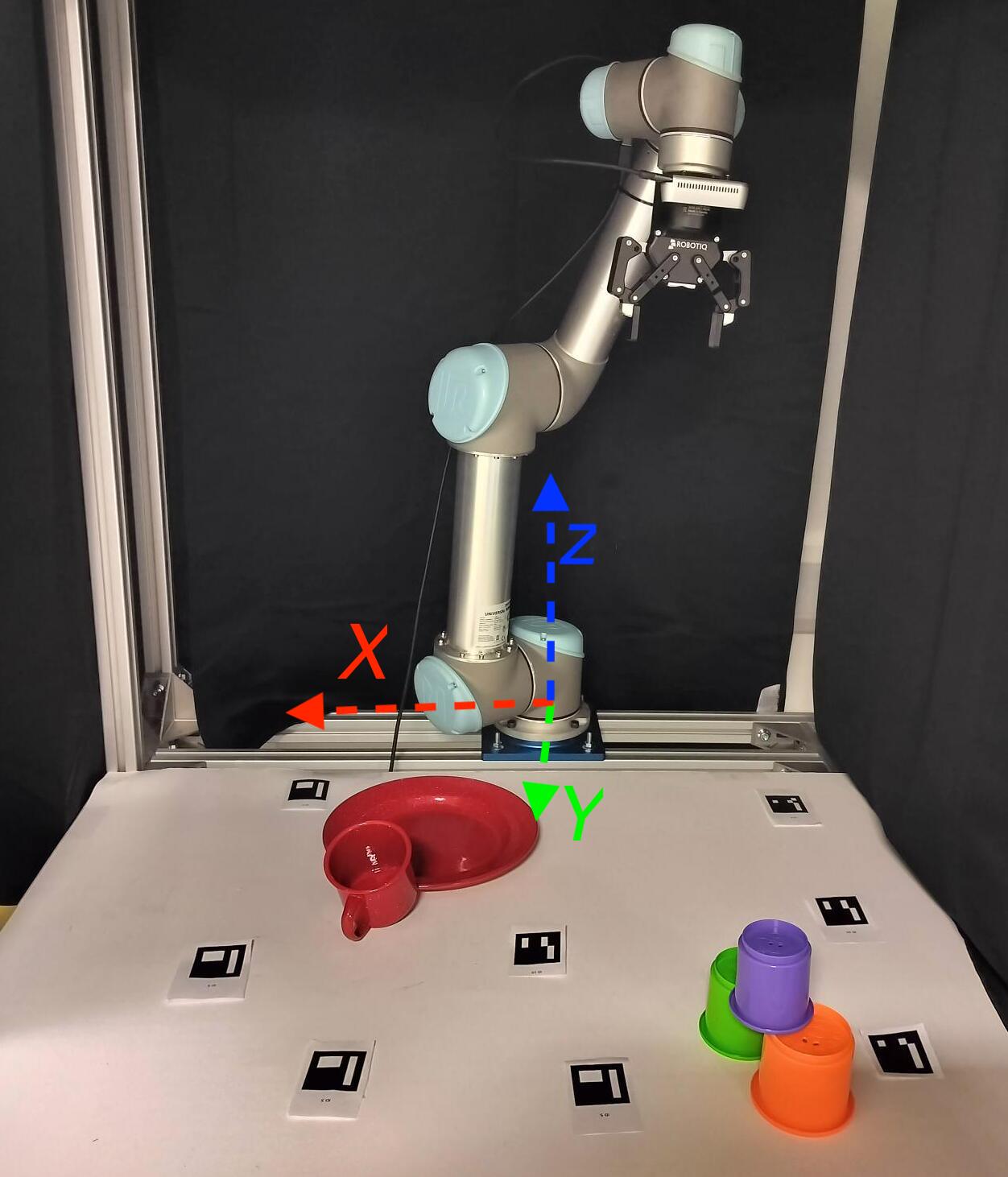}} 
        \vspace{-.5cm}
        \label{fig:scene2}
    \end{subfigure}
    \hfill
    \begin{subfigure}[b]{0.48\linewidth}
        \centering
        \vspace{.3cm}
        {\includegraphics[width=\linewidth,trim={0cm 0cm 0cm 0cm},clip]{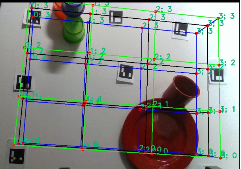}}  
        \vspace{-.5cm}
        \label{fig:overhead_grid}
    \end{subfigure}
    \hfill
    \begin{subfigure}[b]{0.48\linewidth}
        \centering 
        \vspace{.3cm}
        {\includegraphics[width=\linewidth,trim={0cm 0cm 0cm 0cm},clip]{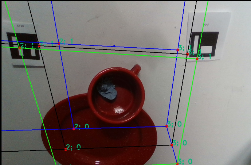}}  
        \vspace{-.5cm}
        \label{fig:overhead_grid}
    \end{subfigure}
    \caption{Overview of the robot setup and grid projection. (Top left and right) The robot in the home configuration for Scene 1 and Scene 2. The virtual 3D grid overlaid on the table projected onto the 2D image for Scene 2, the top-view (bottom left) and goal-view (bottom right) of the target in Scene 2.}
    \label{fig:combined_figures}
    \vspace{-0.5cm}
\end{figure}

The contribution of the paper includes demonstrating how to take advantage of the reasoning and visual understanding abilities of  VLMs for robot active perception. This is achieved via the proposed AP-VLM framework, especially  the overlaid virtual 3D grid, which enables the VLM to connect what it understands to  precise positions for  the robot.
This enhances robot perception and bridges the gap between high-level reasoning and real-time control, advancing the field of Active Perception for information gathering in robotics. 

\section{Related Works}

Research has made significant progress in addressing the limitations of large models in spatial and physical reasoning.
SpatialVLM \cite{chen2024spatialvlm} suggests that VLMs' difficulty with spatial reasoning stems from insufficient spatial training data. Improving spatial representation could enhance tasks that require better object relationship understanding. NEWTON \cite{wang2023newton} explores physical reasoning in LLMs by defining attributes such as malleability and stiffness. Their 160K QA dataset enhances physical reasoning across these attributes. Similarly, PROST \cite{aroca2021prost} tests physics-based commonsense reasoning by asking LLMs to determine object interactions, like identifying stackable items. PIQA \cite{bisk2020piqa} focuses on affordance reasoning, challenging models to understand object properties and affordances.

Recent works have explored LLM integration with robotic manipulation. SayCan \cite{saycan2022arxiv} computes affordance functions from natural language instructions, while Code as Policies \cite{liang2023code} translates LLM-generated code into executable robot policies. ProgPrompt \cite{singh2023progprompt} generates Pythonic headers to map actions to environment objects. Tidybot \cite{wu2023tidybot} uses in-context learning, prompting LLMs with human-derived object placement preferences and leveraging VLMs for enhanced detection.

Visual prompting has also addressed robotic perception challenges. MOKA \cite{liu2024mokaopenworldroboticmanipulation} uses VLMs to predict scene affordances by annotating images with grasp and target points. In contrast, our approach extends the grid to three dimensions to provide direct end-effector locations. PIVOT \cite{nasiriany2024pivot} annotates images with numbered keypoints, enabling robots to capture additional images and refine scene understanding, aligning with our active perception framework.

Active perception plays a crucial role in enhancing robotic performance in complex environments. Kwon et al. \cite{kwon2024groundedcommonsensereasoning} use active perception to select alternative views when the current view is insufficient, relying on a continuous QA loop between a VLM and LLM. However, their approach depends on pre-collected images, some of which come from unreachable viewpoints. Active perception reduces epistemic uncertainty, as emphasized by Kroemer \cite{kroemer2021review}, who contrasts passive and active modes. Celemin et al. \cite{celemin2023knowledge} highlight that passive perception often leads to ambiguity due to insufficient sensory input.

The complexity of active perception, as discussed by Li et al. \cite{li2023internally}, has limited its use despite its advantages. Recent research has explored multimodal data integration for improved perception, though this approach faces challenges like numerical instability and generalization issues \cite{wang2022uncertainty}. Active information gathering, as emphasized by Bohg et al. \cite{bohg2017interactive}, helps reduce uncertainty and aids decision-making. Hierarchical planning models \cite{sutton1999between, sharma2021skill} are integral to our framework, where LLMs and VLMs guide the robot's movements iteratively.

\begin{figure*}[htb]
    \centering
    \includegraphics[width=\textwidth]{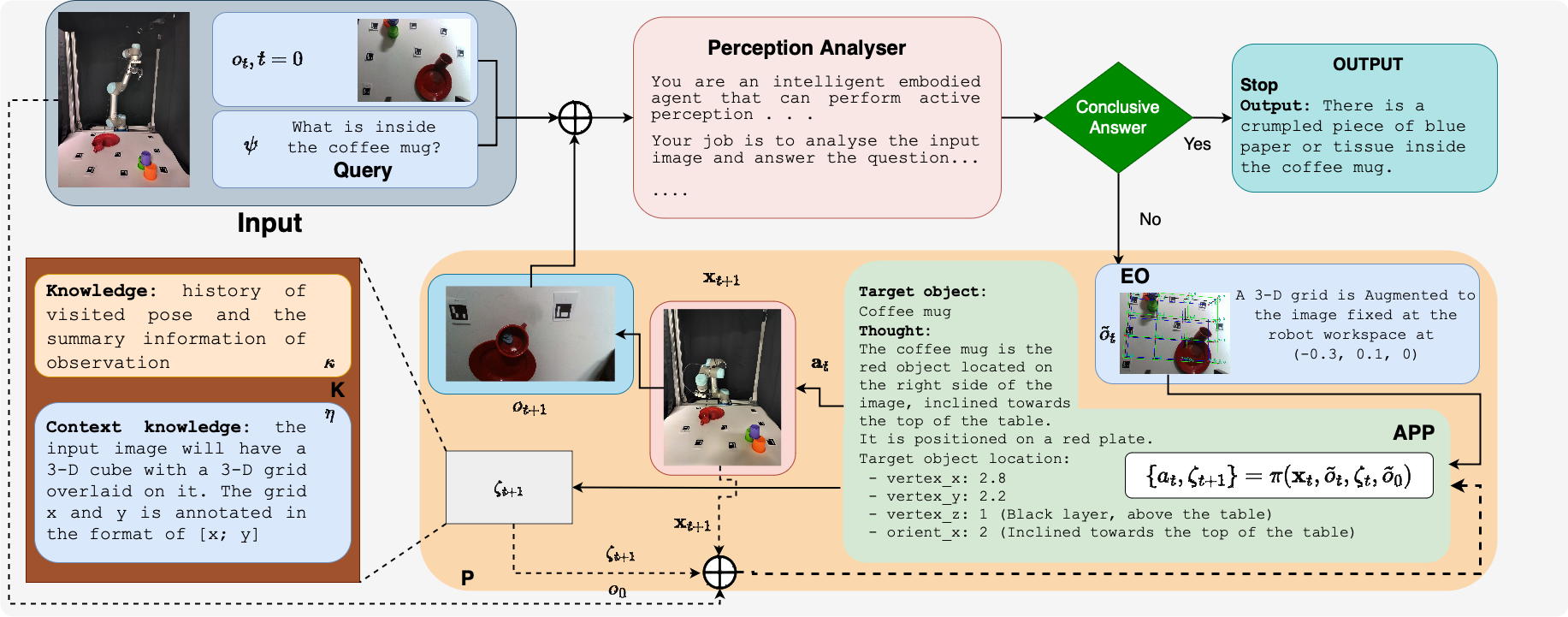}
    \caption{\textbf{Schematic of the Proposed Active Perception via Vision-Language Model (AP-VLM) Framework:} The framework consists of four main components: (i) Input, (ii) Perception Analyzer, (iii) Output, and (iv) Policy Block (\textbf{P}). The Perception Analyzer first attempts to answer the query. If unsuccessful, the Enhanced Observation (EO) block overlays a 3D virtual grid onto the input image, anchored at the robot’s workspace origin $(-0.3, 0.1, 0)$, and passes it to the VLM agent. The Active Perception Policy (APP) then uses the augmented image ($\tilde{o}_t$) from the EO and the home configuration image ($o_0$) to generate an action $a_t$ through the policy $\pi(a_t, \zeta_{t+1} | \mathbf{x}_t, \tilde{o}_t, \zeta_t, o_0)$ and updates the knowledge $\zeta_{t+1} = \{\eta_0, \kappa_{t+1}\}$, where $\eta_0$ represents the context knowledge from the initial prompt, and $\kappa_{t+1}$ represents the knowledge obtained at each control iteration. After executing the action, the robot transitions to a new state $\mathbf{x}_{t+1}$ and captures a new observation ($o_{t+1}$). The Perception Analyzer processes this new image and attempts to answer the query again. The process repeats until the Perception Analyzer provides a satisfactory answer or after a maximum of 10 iterations.}
    \label{fig:method_RL}
\end{figure*}
\section{Methodology}

In this work, we approach the problem of active perception for a robotic manipulator integrated with a VLM. The system, consisting of a robotic manipulator equipped with an in-hand camera (Fig.~\ref{fig:combined_figures}), acts as an \textit{agent} that interacts with its environment to maximize information gain and answer specific queries. The task involves analyzing the scene and moving the robot to different viewpoints in a virtual coordinate frame $F$ (detailed in the Enhanced Observation part) to iteratively gather visual data until the VLM can provide a conclusive answer. An overview of this framework is shown in Fig.~\ref{fig:method_RL}.

\noindent \textbf{Problem Formulation:} Given a natural language query and visual observation $o$ at time $t$, the agent uses a \textit{Perception Analyzer} to output a binary result $Z$. Based on this output, the agent decides whether to terminate or generate a new action using the Active Perception Policy (APP) $\pi$. This policy is defined based on the agent’s current state $s_t$, which consists of the robot’s current pose $\textbf{x} = (p, \theta)$, the \emph{Query} $\psi$. and the context $\zeta = \{\eta,  \kappa\}$, where $\eta$ is the agent’s initial knowledge, which includes a set of principles guiding the agent's actions (such as action constraints and goals), and $\kappa$ is the cumulative knowledge obtained during active perception process (e.g. visited vertices during the past active perception iterations), provided to the VLM.  
The generated action is mapped from the camera coordinate frame $C$ to the robot's base frame $B$, enabling the robot to collect new observations. To facilitate 3D scene understanding for the VLM, we propose using a virtual cube projected onto the 2D image space.

\noindent \textbf{Enhanced Observation:} We introduce the Enhanced Observation (EO) (Fig.~\ref{fig:combined_figures}, bottom left), which enhances the VLM's ability to generate policies by incorporating a parameterizable 3D grid into the agent's observation. At each time step, the agent can provide spatial information along all three axes using ArUco markers for depth estimation relative to the robot's base frame. A reference ArUco marker $M$ is chosen, and vertices $v$ are defined around it in $C$, denoted as $V^{M}_{v}$. These vertices are projected onto the table surface, forming a virtual cube divided into grids.

To transform these vertices into $B$, we compute the corresponding homogeneous transformation matrix $T$:

\[
V_{v}^{B} = T_{C}^{B} T_{M_{id}}^{C} V_{v}^{M_{id}}
\]

Here, $T_{M_{id}}^{C}$ is derived from the camera's intrinsic parameters using translation and rotation vectors, and $T_{C}^{B}$ is generated using the robot's inverse kinematics. These transformations ensure the 3D grid remains aligned with the robot's workspace.

The translation and rotation vectors are continuously updated as the robot moves, maintaining a consistent 3D grid projection despite the robot's changing position. Multiple ArUco markers $M_{id}$ ensure that at least one marker is visible to the camera at all times, allowing for continuous and accurate 3D grid generation and projection onto the 2D image.

\noindent Each vertex in the 3D grid is annotated along the $x$ and $y$ axes in the format $(x; y)$ to aid spatial interpretation.

\noindent \textbf{State $(S)$}: The state $s_t$ at time $t$ represents the robot's current configuration and the surrounding scene. It includes the robot’s position, orientation, and relevant spatial information:
\[
s_t = \{ \{\textbf{x}_1, \zeta_1 \}, \dots, \{\textbf{x}_N, \zeta_N \} \}
\]

\noindent \textbf{Observation $(O)$}: The robot’s observation $o_t$ consists of the camera image capturing the scene, providing real-time visual feedback for perception and task execution.

\noindent \textbf{Active Perception Policy $(\pi)$}: The Active Perception Policy (\textbf{APP}) governs decision-making based on the current observation $o(s_t)$ and an overlay image $\tilde{o}_t$ that includes a 3D grid superimposed on the scene. Anchored at a reference marker, this grid helps the VLM suggest the next optimal action (grid vertex) for gathering further information. The APP balances exploration and task objectives by guiding the robot to actions that maximize the likelihood of answering the query.

\noindent \textbf{Action $(A)$}: The action space $A$ consists of discrete movements, where the robot transitions between configurations $q_t$ and $q_{t+1}$. Starting from the home configuration $o_0 = (-0.1, 0.3, 0.8)$, we evaluate eight action spaces to assess their impact on AP-VLM's performance:
\begin{itemize}
    \item \textbf{NAP}: No active perception (baseline).
    \item \textbf{2DNA}: Movements to vertices on a 2D grid spaced 0.2 m apart and located 0.1 m above the table.
    \item \textbf{2DA}: Movements to vertices described in 2DNA, but with annotated vertex coordinates in the camera frame.
    \item \textbf{3DD}: 3D discrete, moves to vertices on  3D grid (0.6 m x 0.6 m x 0.3 m) with grid heights separated by 0.1 m.
    \item \textbf{3DC}: 3D continuous, movements to any point within the virtual cube described in 3DD.
    \item \textbf{3Dx}: Movements to points in 3DC with 35-degree rotations around the robot’s $x$ axis.
    \item \textbf{3DxN}: Movements to points in 3Dx without considering initial observation $o_0$ in VLM’s context.
    \item \textbf{3Dxy}: Movements to points in 3Dx with 35-degree rotations around the $y$ axis.
    
\end{itemize}

Each action modifies the robot’s camera view, allowing it to gather new perspectives of the scene. Grid vertices are mapped to pre-calibrated coordinates in the robot’s base frame, ensuring precise movements within the workspace.

\subsection{System Architecture}

The system (Fig.~\ref{fig:method_RL}) combines a VLM and a robotic manipulator to capture observations from the scene and answer a specified \emph{Query} $\psi$.

\noindent \textbf{Perception Analyzer:} The VLM acts as the Perception Analyzer, interpreting the visual data captured by the robot’s camera. Its role is to determine whether the current observation is sufficient to answer the query. At each time step, the Perception Analyzer evaluates the scene for relevant objects, spatial relationships, and context, outputting a binary signal. If the observation is conclusive, the process terminates; otherwise, further exploration is triggered through the Active Perception Policy (APP).

\noindent \textbf{Enhanced Observation (EO):} The EO component overlays a 3D grid onto the visual data, aligned with the robot’s workspace. This grid provides structured guidance for the robot's exploration, where each vertex represents a potential new viewpoint to improve scene understanding.

\noindent \textbf{Active Perception Policy (APP):} The APP, aided by the EO grid, directs the robot’s movements to gather additional information when the current observation is insufficient. For tasks like identifying an object inside a mug, the grid helps the APP determine the best positions and orientations $\mathbf{x}_{t+1}$ for further exploration. The grid coordinates are mapped to the camera frame, ensuring precise and repeatable actions.

\noindent \textbf{Knowledge:} The Knowledge component stores both the initial context ($\kappa$) from the task prompt and the history of observations and grid vertices ($\eta$). This prevents the robot from revisiting previously explored positions and allows the system to refine its understanding of the scene. 
The knowledge repository ensures efficient exploration and reduces redundancy, optimizing the exploration process.

\noindent \textbf{Input Block:} The Input Block processes the initial visual data from the robot’s camera at its home configuration and the natural language query (e.g., "What is inside the mug?"). This provides the system with the starting conditions for exploration, including the robot’s pose and the task query. From this point, the VLM acts as the Perception Analyzer, guiding the robot's actions through the APP.

\noindent \textbf{Iterative Exploration Process:} The robot starts at an initial state $s_0$ in its home configuration, capturing an image of the scene annotated with a virtual grid. The VLM evaluates the image to determine if sufficient information is available. If not, the VLM suggests a new viewpoint with reference to the grid, and the robot moves to that position. At each subsequent state $s_{t+1}$, the robot captures a new image, and the VLM is queried again. 

\noindent \textbf{Termination Criteria:} The episode ends when (1) the VLM confidently answers the query with a high certainty score, or (2) the system reaches a predefined limit on the number of actions, indicating no further improvement is possible. 

\section{Experiment Setup and Structure}

Here we detail the hardware and Vision-Language Model (VLM) configurations used in our experiments. We evaluate the performance of our Active Perception framework across various exploration strategies and scenarios. The experiments were conducted on two distinct environments, each tested with multiple configurations to assess the system's effectiveness in accurately resolving queries.

\noindent \textbf{Robotic Setup:}  
For our active perception experiments, we used two robotic arms: a 6-DOF UR5 robotic manipulator for Experiment set 1 and a 7-DOF Franka Emika robotic arm for Experiment set 2. Both robotic arms were equipped with an RGB-D camera for capturing visual data from the environment and were used to perform actions such as movement across a virtual 3D grid and orientation adjustments. We use selected objects from the \emph{YCB} object set~\cite{calli2015ycb}.

We used an Intel RealSense RGB-D camera mounted on the end-effector of both robotic arms. The camera provided real-time RGB data for assessing the spatial relationships within the scene and capturing observations at various viewpoints as part of the iterative exploration process (we did not use depth). To ensure precise localization and movement, we used 9 \emph{ArUco markers} strategically placed around the workspace. These markers anchored the 3D grid overlaid on the robot’s environment, providing reference points for accurate positioning of the robotic arms, ensuring consistency in capturing visual data from different perspectives.

For our VLM, we employed the GPT-4o model, which served as the perception engine responsible for interpreting the visual data and guiding the robot’s actions. The VLM processed input from the camera and generated responses to natural language queries, such as "What is inside the mug?". Additionally, the model suggested optimal viewpoints for the robot to explore when further information was required.

GPT-4o was integrated into the system as both the Perception Analyzer and decision-making component, working alongside the Active Perception Policy (APP). It analyzed the visual input from the camera and assessed whether the current scene provided sufficient information to resolve the query. If not, the VLM provided feedback to the APP, guiding the robot to a new position within the workspace. The VLM’s capability to suggest spatially informed actions, such as moving to specific grid locations, enabled a more intelligent exploration process.



The experiments on the eight action spaces were structured to emphasize the importance of various elements of the AP-VLM framework. Specifically, we evaluated:
\begin{itemize}
    \item (\textbf{i}) The impact of active perception (AP) compared to a baseline scenario without active perception.
    \item (\textbf{ii}) Using a 2D or 3D grid, with and without annotation, to assess how dimensionality and visual aids affect the robot's ability to gather relevant information.
    \item (\textbf{iii}) Using discrete or continuous vertex outputs from the VLM in guiding the robot's movements.
    \item (\textbf{iv}) Allowing rotations around the $x$, or $x$ and $y$. 
\end{itemize}

Each scene was analyzed 10 times under each action space resulting in 160 total trials, and key metrics were recorded to evaluate the system’s performance.

We evaluated the framework based on metrics adapted from the R2R-VLN dataset \cite{anderson2018vision}, modified for use in our manipulation setup:
\begin{itemize}
    \item \emph{Success Rate} (\textbf{SR}): The percentage of times the agent correctly answered the input query.
    \item \emph{Total Length (Position)} (\textbf{TLP}): The mean total distance travelled by the robot across all trials.
    \item \emph{Total Length (Position, Successful)} (\textbf{TLPS}): The mean total distance traveled by the robot across successful trials.
    \item \emph{Position Error} (\textbf{PE}): The mean Euclidean distance between the robot's final position and the target position.
    \item \emph{Orientation Error} (\textbf{OE}): The mean quaternion error between the robot's final orientation and the target orientation.
    \item \emph{Oracle Success Rate} (\textbf{OSR}): The success rate when the agent stops at the closest point to the goal along its trajectory. A margin of error of $0.1\: m$ was allowed.
\end{itemize}

\begin{table*}[t!]
\centering
\caption{Performance Metrics for Scene 1 and Scene 2 for our models.}
\label{tab:results}
\begin{tabular}{|l|c|c|c|c||c|c|c|c|}
\hline
\multirow{2}{*}{\textbf{Method}} & \multicolumn{4}{c||}{\textbf{Scene 1}} & \multicolumn{4}{c|}{\textbf{Scene 2}} \\ \cline{2-9}
 & \textbf{SR $\uparrow$} &\textbf{TLP, TLPS $\downarrow m$} & \textbf{PE, OE} $\downarrow m, deg$ & \textbf{OSR $\uparrow$} & \textbf{SR $\uparrow$} & \textbf{TLP,TLPS $\downarrow m$} & \textbf{PE, OE $\downarrow m, deg$} & \textbf{OSR} $\uparrow$ \\ \hline
NAP              & 0.0         & \{\textbf{0.82}, \textbf{0.0}\}           & \{0.21, \; - -\}         & 0         & 0.0         & \{\textbf{0.83}, 0.0\}         & \{0.26,  \quad - -\}         & 0        \\ \hline
2DA              & \textbf{1.0}         & \{1.09, 1.09\}          & \{0.13, \; - -\}         & 9         & 0.2         & \{1.98, 2.05\}         & \{0.30,  \quad - -\}         & 4         \\ \hline
2DNA             & 1.0         & \{0.78, 0.78\}          & \{0.129, \; - -\}        & 0         & 0.2         & \{2.33, 3.51\}         & \{0.44,  \quad - -\}         & 1         \\ \hline
3DD              & 0.8         & \{1.36, 1.11\}          & \{0.173, \; - -\}        & 3         & 0.3         & \{2.21, 1.94\}         & \{0.21,  \quad - -\}         & 3         \\ \hline
3DC              & 0.9         & \{1.00, 0.93\}          & \{0.155, \; - -\}        & 2         & 0.2         & \{2.39, 2.00\}         & \{0.26, \quad - -\}         & 6         \\ \hline
3Dx             & 0.8         & \{0.948, 0.85\}         & \{0.09, 82.78\}     & 10        & \textbf{0.5}         & \{{1.30}, 1.13\}         & \{\textbf{0.15}, \textbf{39.92}\}     & \textbf{8}        \\ \hline
3DxN             & 0.8         & \{0.956, 0.95\}         & \{\textbf{0.08}, \textbf{56.22}\}     & \textbf{11}        & 0.4         & \{1.46, 1.14\}         & \{0.22, 80.46\}     & 4         \\ \hline
3Dxy             & {0.5}         & \{0.961, 0.81\}         & \{0.138, 85.97\}    & 9         & 0.1         & \{1.68, 0.98\}         & \{0.19, 133.66\}    & \textbf{8}         \\ \hline
\end{tabular}
\end{table*}

\section{Results and Discussion}

We demonstrate the cross-functionality of the AP-VLM framework using two robotic manipulators with varying degrees of freedom: a 6-DOF UR5 (Exp1) and a 7-DOF Franka Panda (Exp2) arm.

\noindent \textbf{Exp1:} Table~\ref{tab:results} summarizes the performance metrics for Scene 1, where the object is perpendicular to the surface, and Scene 2, where the object is inclined towards the top of the table (Fig. \ref{fig:combined_figures}). The results clearly highlight the effectiveness of active perception in enhancing the robot's understanding of the scene.

\noindent \textbf{Scene 1:}  
In Scene 1, where the object (aluminium tin) is placed perpendicular to the surface, active perception methods significantly improve the Success Rate (SR). The SR reaches 1.0 with both the 2DA and 2DNA methods, which employ a 2D grid with annotated vertices. These results suggest that even basic 2D grids can provide sufficient spatial understanding in simpler environments where objects are relatively easy to perceive.

However, methods using a 3D grid (such as 3Dx and 3DxN) outperform the 2D methods in reducing Position Error (PE) and Orientation Error (OE), particularly when orientation adjustments are incorporated. For instance, the 3Dx method achieves the lowest PE of 0.09 m and introduces an orientation adjustment that reduces OE to 82.78 degrees. This demonstrates the value of integrating 3D spatial information and orientation handling to enhance the robot's ability to accurately resolve the query. The NAP (No Active Perception) baseline fails to achieve any success, with an SR of 0 and a relatively high TLP of 0.825 m, indicating that passive observation alone is insufficient for task completion in this scenario.

\noindent \textbf{Scene 2:}  
Scene 2 presents a more challenging environment with the object (coffee mug) inclined towards the table. The increased complexity causes the Success Rate (SR) to drop significantly for most methods, with only the 3Dx method achieving a relatively high SR of 0.5. 

Orientation-enhanced methods, such as 3Dx and 3DxN, show clear advantages in Scene 2. Specifically, 3Dx records an OE of 39.92 degrees, the lowest in this scene, along with a PE of 0.15 m. These results suggest that orientation adjustments are crucial for improving performance in complex environments, where the robot must view the scene from different angles to gain a better understanding. The OSR metric shows that even when the SR is low, methods like 3Dxy and 3Dx maintain higher OSR values, indicating that the robot was close to solving the query but did not complete the task within the given constraints.

\begin{figure}[tb!]
    \centering
    \includegraphics[width=0.6\linewidth,trim={10cm 0cm 12cm 0cm},clip]{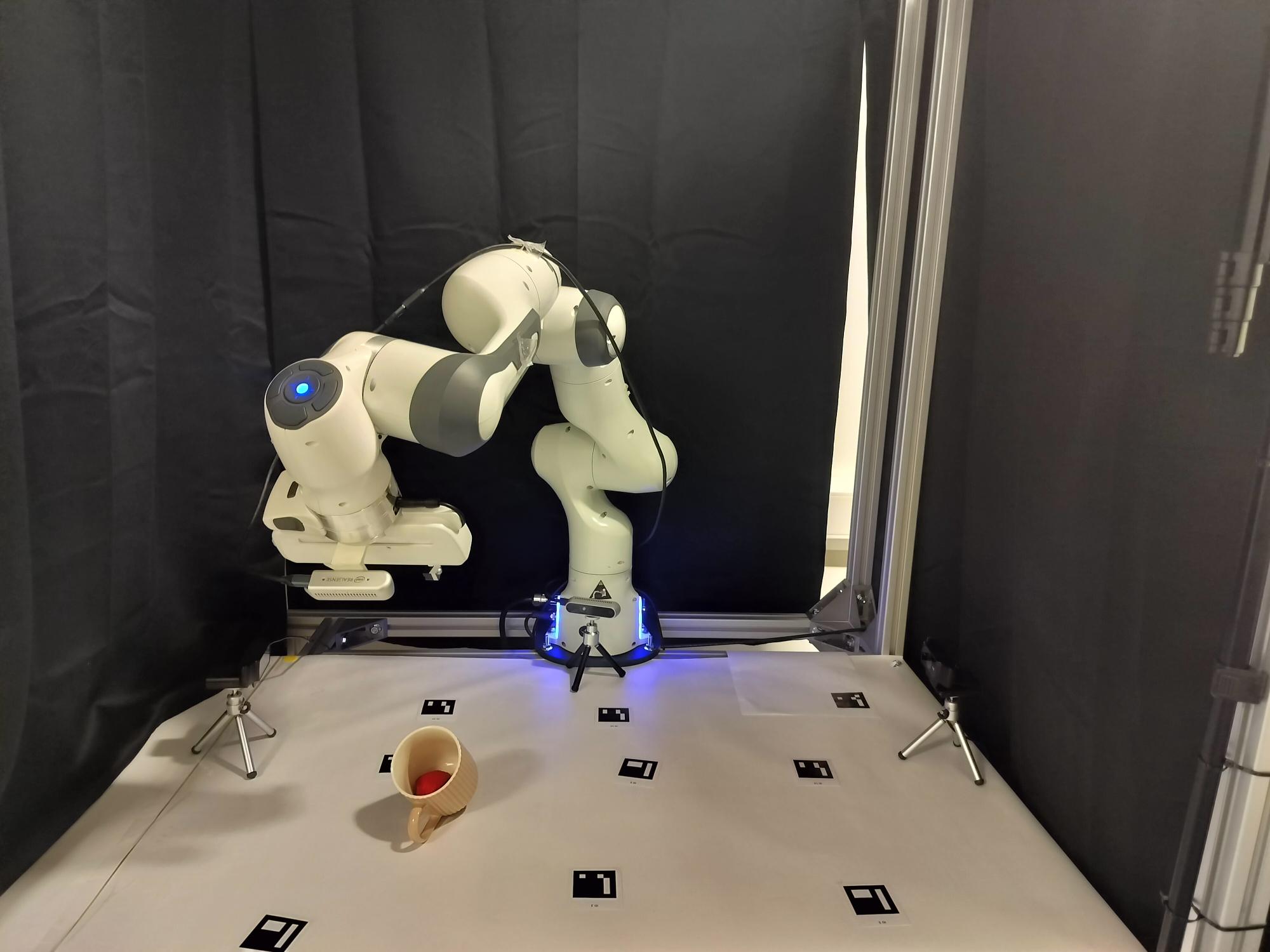}
    \caption{Robotic set up 2 demonstrates AP-VLM is robot agnostic. A 7-DOF Franka Panda arm is used for comparison. AP-VLM moved the arm to the goal pose to look into the mug. }
    \label{fig:expsetup2}
    \vspace{-0.5cm}
\end{figure}
In both scenes, it is evident that active perception (AP) methods outperform passive perception (NAP). The use of a 3D grid (3Dx, 3DxN, 3Dxy) consistently yields more accurate results compared to 2D grid methods, especially when orientation adjustments are incorporated. Moreover, the higher TLP in methods like 2DNA and 3DD is because  these methods  explore a larger space;  they tend to go around the target and not on the target where they can get the best view to give an answer. They are therefore less efficient in refining the robot's perception compared to orientation-aware methods like 3Dx and 3DxN. 
This highlights the importance of optimizing both position and orientation during the exploration process.
Nevertheless, the VLM occasionally struggles to handle combined orientation actions around both the x and y axes; the VLM  adds x or y-rotations when both or at least one of them are not required. The VLM seems to get more  confused with multiple options. 

\begin{figure*}[tb!]
    \centering
    \begin{subfigure}[b]{0.3\linewidth}
        \centering
        \includegraphics[width=\linewidth, trim={11cm 0 9cm 0}, clip]{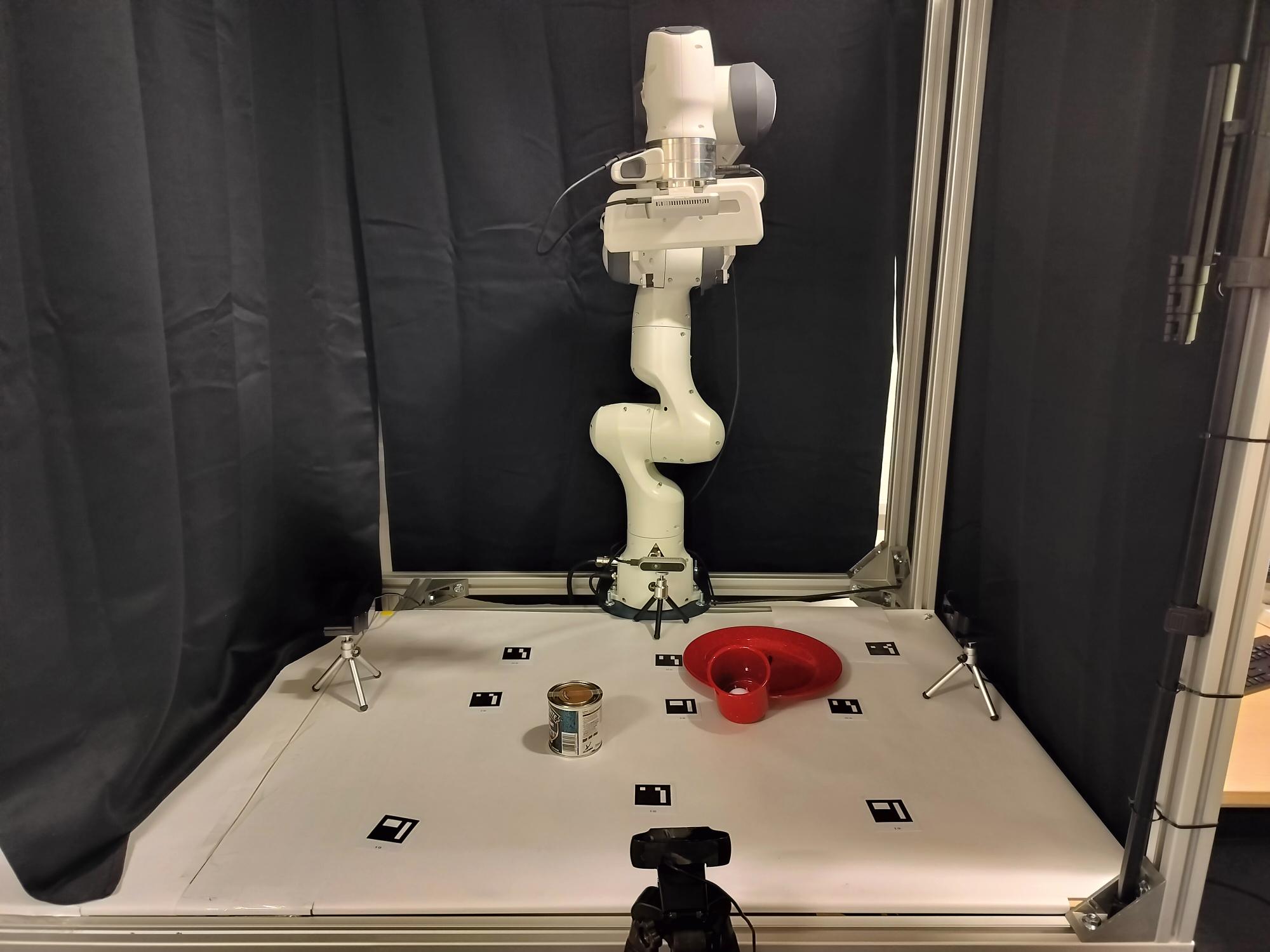}  
        \caption{}
        \label{fig:exp2}
    \end{subfigure}
    \hfill
    \begin{subfigure}[b]{0.3\linewidth}
        \centering
        \includegraphics[width=\linewidth, trim={11cm 0 9cm 0}, clip]{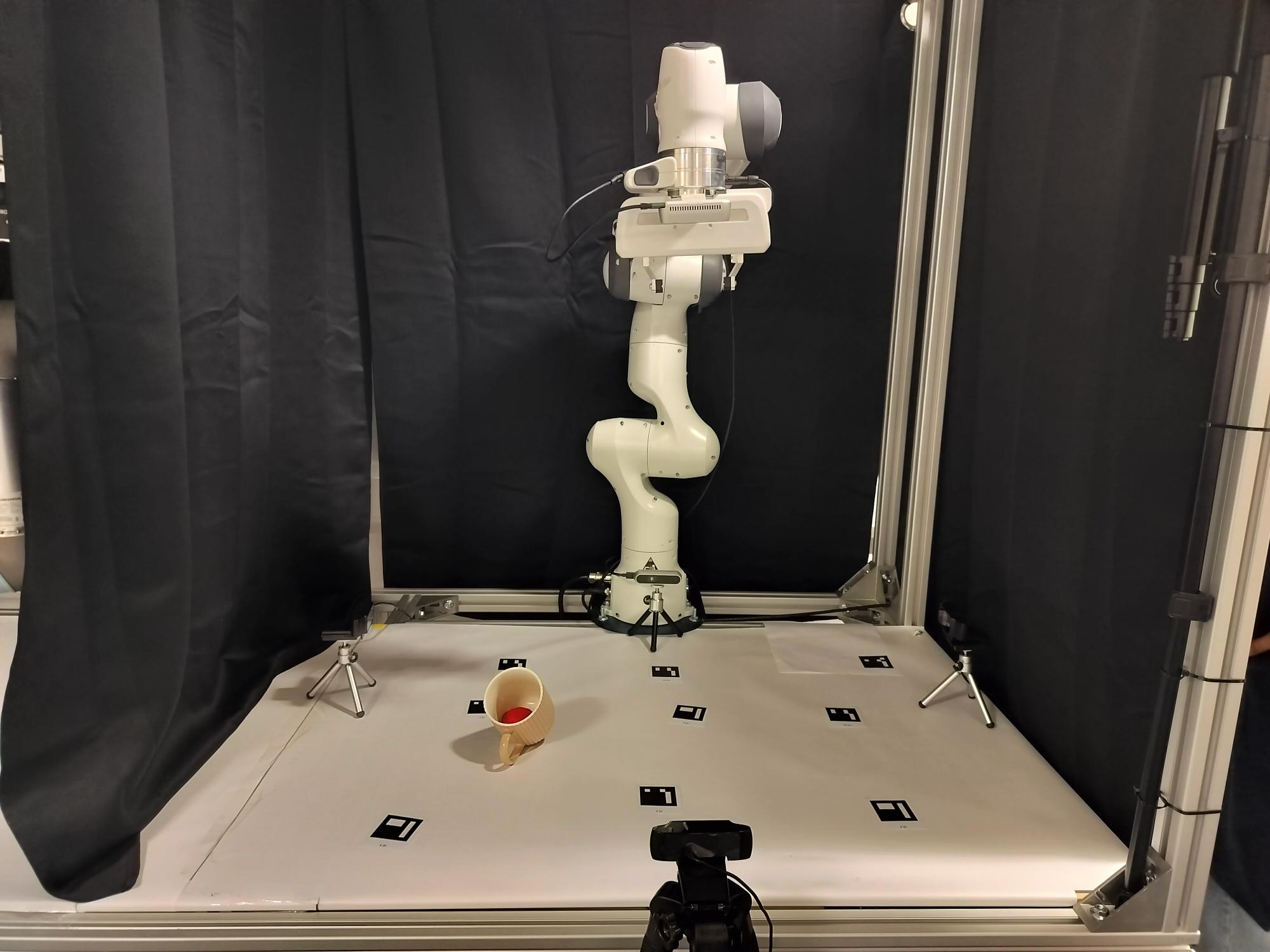}  
        \caption{}
        \label{fig:exp3}
    \end{subfigure}
    \hfill    
    \begin{subfigure}[b]{0.3\linewidth}
        \centering
        \includegraphics[width=\linewidth, trim={11cm 0 9cm 0}, clip]{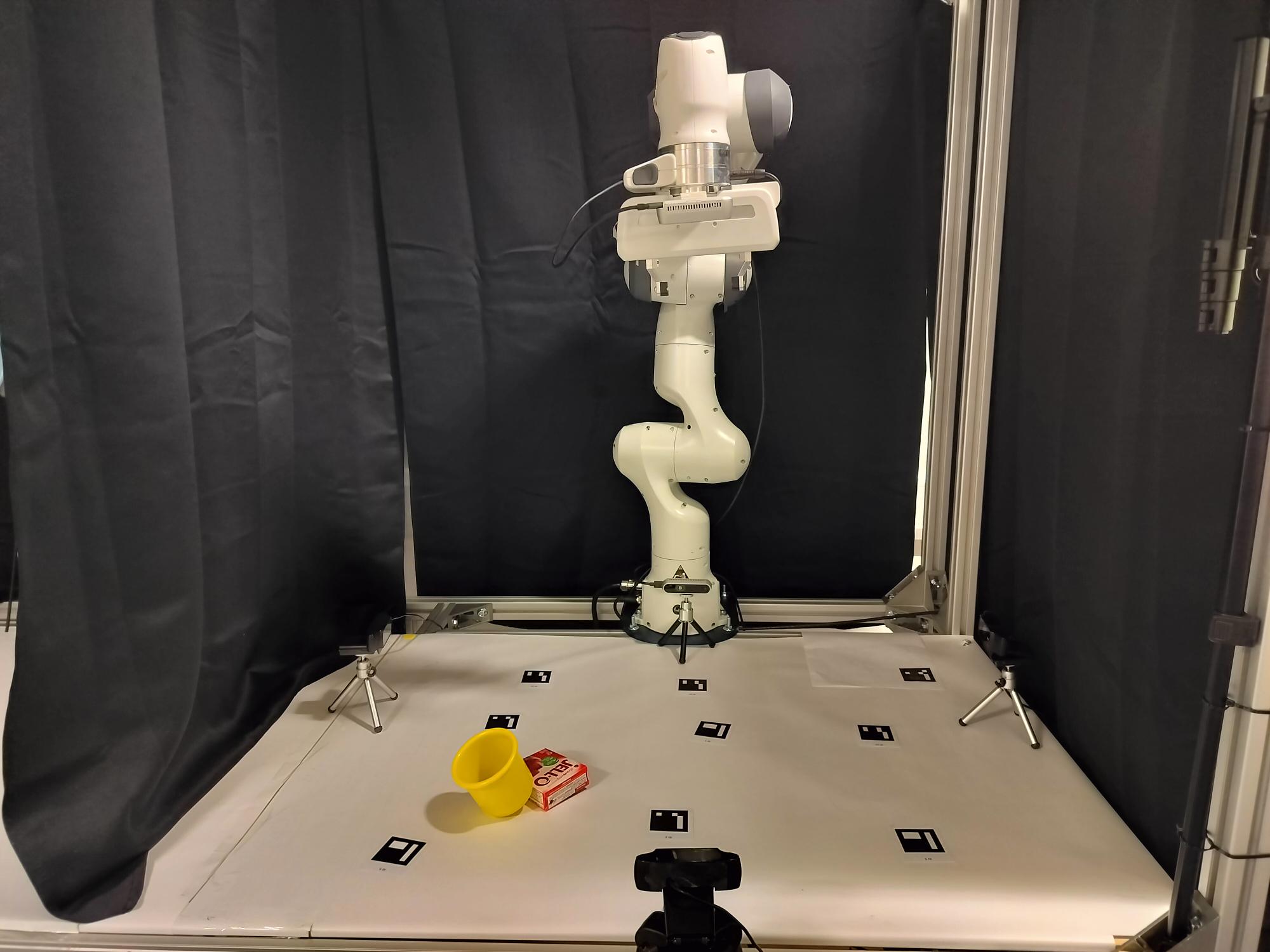}  
        \caption{}
        \label{fig:exp1}
    \end{subfigure}
    \caption{Scenes used for comparing AP-VLM with TGCSR: (a) Scene 1:  a golf ball inside a coffee mug; (b) Scene 2: a plastic strawberry inside a cup; and (c) Scene 3: a Lego block inside a cup inclined close to the table surface.}
    \label{fig:experiments}
\end{figure*}

\noindent \textbf{Comparison with Baseline (Exp2):}  
We further evaluated our AP-VLM framework by comparing it with the Toward Grounded Common Sense Reasoning (TGCSR)~\cite{kwon2024groundedcommonsensereasoning} model using the Franka Panda robot in three scenes. We tested our best-performing model (3Dx) against TGCSR, which requires discrete images to be collected, whereas AP-VLM dynamically gathers observations based on the task requirements. For TGCSR, we positioned fixed cameras at five distinct locations, 15 cm above the table surface (front, back, sides view) and at top view, $(-0.1, 0.3, 0.8)\; m$, to collect images (Fig.~\ref{fig:expsetup2}).

The success rates were evaluated over five trials per scene for each model, resulting in 30 trials in total. Overall, AP-VLM achieved a higher success rate since it was not restricted to discrete camera views. We considered 3 scene in this comparison study (Fig.~\ref{fig:experiments}). In Scene 1, where a golf ball inside a coffee mug was partially visible, TGCSR misinterpreted the scene and incorrectly concluded that the mug was filled with liquid (0/5  successful trials), while AP-VLM successfully moved closer and correctly identified the object (5/5 successful trials). In Scene 2, where a strawberry inside a cup was completely hidden from fixed camera views, TGCSR failed in all trials (0/5 successful trials), while AP-VLM succeeded by adjusting its position and orientation to capture a clear view inside the cup (5/5 successful trials). The third scene, involving a cup inclined close to the table surface with a Lego block inside, was challenging for both models. AP-VLM struggled to find an optimal viewing angle (1/5 successful trials) but still outperformed TGCSR, which failed entirely (0/5 successful trials).

\noindent \textbf{Discussion:}  
The results from both experiments confirm the advantages of incorporating active perception, 3D grids, and orientation adjustments for resolving queries in complex environments. In Scene 1, where the object is perpendicular to the table, simpler methods such as 2DA and 2DNA can achieve high success rates with minimal orientation adjustments. However, as the complexity increases, as in Scene 2, orientation-aware methods like 3Dx become more effective, significantly improving both positional and orientational accuracy. 

The experiments also underscore the importance of combining spatial and orientational cues in active perception. While simpler environments may suffice with a 2D grid, more complex environments require the integration of 3D grids and orientation adjustments. OSR values further suggest that while some methods approach solving the query, they may require additional iterations or fine-grained orientation adjustments to fully resolve the task.

Lastly, AP-VLM’s ability to dynamically gather observations makes it a superior alternative to systems like TGCSR, which rely on predetermined orientations. This adaptability allows AP-VLM to handle more diverse and complex tasks by intelligently selecting viewpoints and orientations for exploration, leading to more accurate and efficient task completion.

The current AP-VLM framework is limited by discrete orientation adjustments, which hinder the robot's ability to inspect inclined or hollow objects that require precise front views. Sometimes, the actions are out of reachable robot workspace. Future works will focus on solving these. 
\section{Conclusion}

In this paper, we introduced AP-VLM, an active perception framework that integrates vision-language models (VLMs) with robotic systems for dynamic scene exploration. By leveraging active perception, 3D spatial grids, and orientation adjustments, the framework enables robots to iteratively gather visual information to resolve complex queries. Our experiments, conducted with both 6-DOF and 7-DOF robotic arms, demonstrated significant improvements in task success rates, particularly in scenarios where objects were partially occluded or placed in challenging orientations. The adaptability of AP-VLM, compared to baseline methods like TGCSR, highlights the importance of dynamic observation gathering, which allows robots to actively choose optimal viewpoints and orientations during exploration. These results suggest that combining VLMs with active perception can significantly enhance a robot’s understanding of its environment, making it more capable of handling real-world tasks that require both semantic reasoning and precise physical interaction. Future work will focus on extending the framework to more diverse environments and improving its ability to handle more complex object interactions and tasks.

\addtolength{\textheight}{-1cm}   





\bibliographystyle{IEEEtran}
\bibliography{ref}

\end{document}